\documentclass[pdf-a,balance,colorlinks,upint,subscriptcorrection,varvw,mathalfa=cal=boondoxo, spanish,french,vietnamese,russian,greek]{asmeconf}

\hypersetup{%
	pdfauthor={John H. Lienhard},									  
	pdftitle={ASME Conference Paper LaTeX Template},                  
	pdfkeywords={ASME conference paper, LaTeX template, BibTeX style},
	pdfsubject = {Describes the asmeconf LaTeX template},			  
	pdflicenseurl={https://ctan.org/pkg/asmeconf},
}

\begin{document}


\ConfName{Proceedings of the ASME 2024\linebreak International Mechanical Engineering Congress and Exposition}
\ConfAcronym{IMECE2024}
\ConfDate{November 17--21, 2024} 
\ConfCity{Portland, OR} 
\PaperNo{IMECE2024-145673}


\title{Deep Reinforcement Learning for Decentralized Multi-Robot Control: A DQN Approach to Robustness and Information Integration} 
 
%
%
%

\SetAuthors{%
	Bin Wu, 
        C. Steve Suh\CorrespondingAuthor{ssuh@tamu.edu, wubin@tamu.edu} 
	}

\SetAffiliation{}{Texas A\&M University, College Station, TX }



\maketitle

\versionfootnote{Documentation for \texttt{asmeconf.cls}: Version~\versionno, \today.}




\begin{abstract}

The superiority of Multi-Robot Systems (MRS) in various complex environments is unquestionable. However, in complex situations such as search and rescue, environmental monitoring, and automated production, robots are often required to work collaboratively without a central control unit. This necessitates an efficient and robust decentralized control mechanism to process local information and guide the robots' behavior. In this work, we propose a new decentralized controller design method that utilizes the Deep Q-Network (DQN) algorithm from deep reinforcement learning, aimed at improving the integration of local information and robustness of multi-robot systems. The designed controller allows each robot to make decisions independently based on its local observations while enhancing the overall system's collaborative efficiency and adaptability to dynamic environments through a shared learning mechanism. Through testing in simulated environments, we have demonstrated the effectiveness of this controller in improving task execution efficiency, strengthening system fault tolerance, and enhancing adaptability to the environment. Furthermore, we explored the impact of DQN parameter tuning on system performance, providing insights for further optimization of the controller design. Our research not only showcases the potential application of the DQN algorithm in the decentralized control of multi-robot systems but also offers a new perspective on how to enhance the overall performance and robustness of the system through the integration of local information.

\end{abstract}


\section{Introduction}
In the current field of robotics research, the coordinated control of Multi-Robot Systems (MRS) \cite{arai2002advances} has become an important direction, especially in complex scenarios such as search and rescue \cite{queralta2020collaborative}, environmental monitoring \cite{bai2020cooperative}, and automated production. These scenarios often require robots to collaborate without a central control unit, which places higher demands on the efficiency and robustness of decentralized control mechanisms. Traditional robot control systems often rely on centralized control strategies, but in practical applications, this approach often proves inadequate in the face of environmental complexity and dynamic changes \cite{gautam2012review}. Therefore, developing a decentralized controller that can effectively integrate local information and improve system adaptability and fault tolerance is particularly important\cite{arai2002advances, wu2019decentralized}.

To address these challenges, deep reinforcement learning offers a new solution \cite{omidshafiei2017deep,agrawal2023rtaw,kapoor2018multi}. This paper proposes a decentralized controller design method based on the Deep Q-Network (DQN) \cite{fan2020theoretical,ong2015distributed}, aimed at enhancing the efficiency and robustness of information integration in multi-robot systems. Our method allows each robot to make decisions independently based on its observed local state while enhancing the entire system's collaborative efficiency and adaptability to dynamic environment through a shared learning mechanism. Tests in a simulated environment have validated the effectiveness of this controller in improving task execution efficiency, strengthening system fault tolerance, and environmental adaptability. Additionally, this study also explores the impact of DQN parameter adjustments on system performance, providing insights for further optimization of controller design.

This research not only demonstrates the potential application of the DQN algorithm in the control of decentralized multi-robot systems but also offers a new perspective on how to enhance overall system performance and robustness by integrating local information. By experimentally comparing our Communication-Embedded DQN (CE-DQN) algorithm with the standard DQN algorithm across different robot team compositions and task sizes, we further validate the effectiveness and practicality of our approach.


\section{Methods}
\subsection{Problem formulation}
 In decentralized multi-robot systems, considering the decentralized nature of the system, we can use distributed MDP or Decentralized Partially Observable Markov Decision Processes (Dec-POMDP) \cite{guicheng2022review,kraemer2016multi} to describe the problem. Here, we adopt the definition framework of Multi-agent Markov Decision Processes (MAMDP) \cite{choudhury2022scalable,littman1994markov}.

There are \( N \) robots, with the set of robots denoted as \( \mathcal{R} = \{1, 2, ..., N\} \).
The system operates on discrete time steps \( t = 0, 1, 2, ... \).
\( s_t \in \mathcal{S} \) represents the global state of the system at time \( t \), which may include all robots' positions, environmental status, etc.
\( s_t^i \) represents the local state of the \( i \)-th robot at time \( t \). The local state can be the robot's local sensory information, such as the sensed nearby environment.
\( a_t^i \in \mathcal{A}^i \) represents the action taken by the \( i \)-th robot at time \( t \).
\( \mathbf{a}_t = (a_t^1, a_t^2, ..., a_t^N) \in \mathcal{A} \) represents the combined actions of all robots at time \( t \).
Global Transition Function: \( P(s_{t+1} | s_t, \mathbf{a}_t) \) indicates the probability of the system transitioning to state \( s_{t+1} \) under the global state \( s_t \) and joint action \( \mathbf{a}_t \).
\( r_t^i(s_t, a_t^i) \) represents the immediate reward obtained by the \( i \)-th robot based on its local state and action.
\( R(s_t, \mathbf{a}_t) \) represents the immediate reward obtained by the entire system under the state \( s_t \) and joint action \( \mathbf{a}_t \).
\( \pi^i(a_t^i | s_t^i) \) represents the strategy of the \( i \)-th robot for choosing actions based on its local state.
\( \Pi(\mathbf{a}_t | s_t) \) represents the strategy for choosing actions for all robots based on the global state.

In multi-robot systems, the optimization objective is usually to maximize the system's expected cumulative reward. To achieve this, multi-robot systems need to find a joint strategy \( \Pi \) that maximizes the expected cumulative reward starting from the initial state \( s_0 \). This can be expressed as:

\begin{equation}
    V^{\Pi}(s_0) = \mathbb{E} \left[ \sum_{t=0}^\infty \gamma^t R(s_t, \mathbf{a}_t) \mid s_0, \Pi \right]
\end{equation}

\noindent where 
\( s_0 \) is the initial global state.
\( \gamma \) is the discount factor, \( 0 \leq \gamma < 1 \), used to control the weight of future rewards, ensuring the convergence of the cumulative reward.
\( R(s_t, \mathbf{a}_t) \) is the global immediate reward at time step \( t \), given the state \( s_t \) and the joint action \( \mathbf{a}_t \).
\( \mathbb{E} \) represents the expectation, reflecting all possible future states and sequences of actions.

In decentralized control, each robot \( i \) adopts a local strategy \( \pi^i \) based on its local information. The system's joint strategy \( \Pi \) is the combination of all local strategies \( \Pi = (\pi^1, \pi^2, ..., \pi^N) \).

\begin{figure*}[htbp]
\centering
\includegraphics[width=\linewidth]{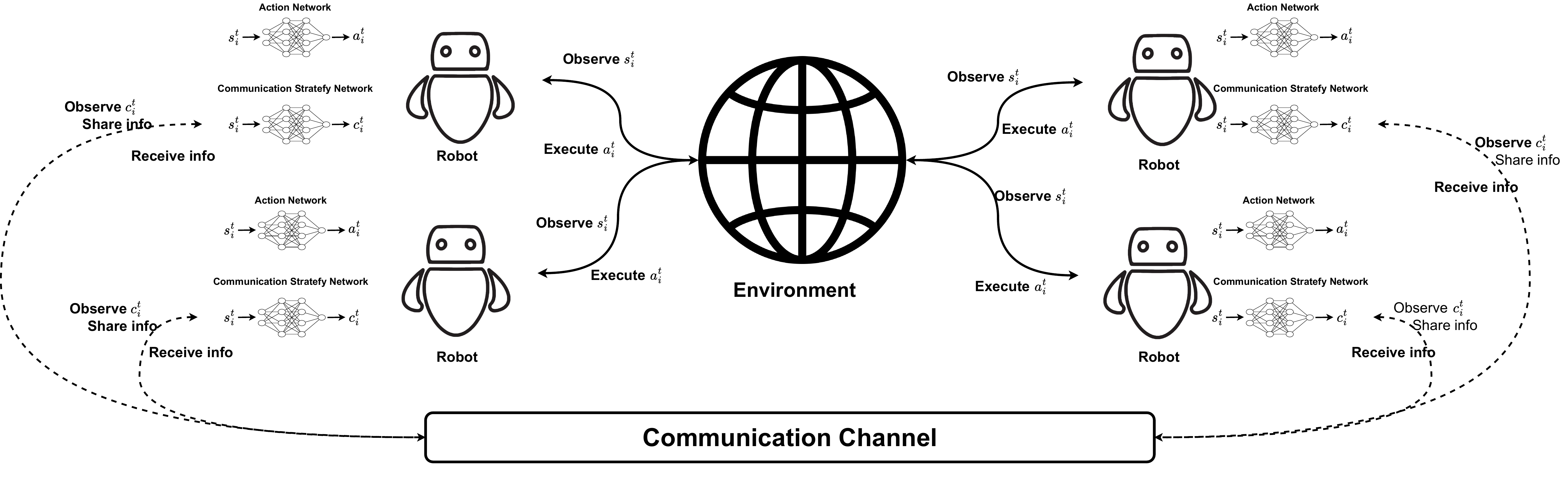}
\caption{Proposed Framework Layer Diagram.}
\label{fig:layer}
\end{figure*}

\subsection{Controller Architecture}
Solving the optimization problem in decentralized multi-robot cooperation presents several challenges, particularly from the perspectives of computational complexity, system scalability, and the utilization of local information.

In multi-robot systems, the dimensions of the state space and action space grow exponentially with the number of robots. The state and possible actions of each robot not only increase the complexity of individual decisions but also rapidly increase the complexity of the problem when considering the joint states and actions of all robots. For each possible state and action combination, it is necessary to calculate the expected reward, which involves traversing all possible subsequent states—a task that is often impractical in reality.

Additionally, in decentralized systems, each robot may only be able to observe limited local information. This limits their ability to make optimal decisions without a global view. The incompleteness of local information makes it more difficult to predict and evaluate global consequences. The optimal decisions of robots may depend on the unknown actions and states of other robots, which complicates the design of strategies based on local information for making independent decisions. In the absence of central coordination, ensuring that local decisions effectively contribute to the global objective is a critical issue.

To address the challenges of controlling decentralized multi-robot systems, we propose a solution based on deep Q-networks (DQN) enhanced with an embedded communication strategy network to facilitate information exchange between robots. This design aims to optimize decision-making quality in decentralized multi-robot systems, improve overall task execution efficiency, and enhance system robustness through effective information sharing. Below, we provide a detailed explanation of our algorithm.

\noindent Step 1: Initialization.
\begin{itemize}
    \item Network Initialization.
    For each robot \( i \), initialize three neural networks:
    Behavior Network \( Q_i \): Used to predict action values.
    Target Network \( \hat{Q}_i \): Aids in stabilizing the learning process and assists in updating the behavior network.
    Communication Strategy Network \( C_i \): Determines when to send or receive information.

    \item Experience Replay Buffer:
    Initialize an experience replay buffer \( D_i \) for each robot \( i \), used to store experience samples including state transitions and rewards received.
\end{itemize}

\noindent Step 2: Execution and Information Exchange
\begin{itemize}
    \item Perform at Each Time Step \( t \):
   Each robot \( i \) selects and executes an action \( a_i^t \) based on the current local state \( s_i^t \) and the behavior network \( Q_i \).
    After executing the action, the robot observes the immediate reward \( r_i^t \) and the new local state \( s_i^{t+1} \).
   Store the transition \( (s_i^t, a_i^t, r_i^t, s_i^{t+1}) \) in the corresponding experience replay buffer \( D_i \).
   \item Decision to Share Information:
   Use the communication strategy network \( C_i \) to evaluate whether the current state \( s_i^t \) is appropriate for sending information:
   \begin{equation}
       c_i^t = \sigma(C_i(s_i^t))
   \end{equation}
    
If \( c_i^t \) exceeds a certain threshold, send key information (such as state, observed events, etc.) to other robots.
Check for information received from other robots and integrate the received information \( b_i \) into their own state representation.
\end{itemize}

\noindent Step 3: Learning Update
\begin{itemize}
    \item Sampling from the Experience Replay Buffer:
    Randomly draw a mini-batch of experiences from the experience replay \( D_i \) for each robot \( i \) for learning purposes.
    \item Calculate Target Q-Values:
        For each sample, use the target network \( \hat{Q}_i \) and the updated state \( s_i' \oplus b_i \) to compute the target values:

        \begin{equation}
             y_i = r_i + \gamma \max_{a'} \hat{Q}_i(s_i' \oplus b_i, a')
        \end{equation}
    \item Update Behavior Network and Communication Strategy Network:
         Update \( Q_i \) by minimizing the prediction error:
     \begin{equation}
         L_i = \left( y_i - Q_i(s_i \oplus b_i, a_i) \right)^2
     \end{equation}
    
        Simultaneously update the communication strategy network \( C_i \), encouraging or discouraging information transmission in specific states.
    \item Periodically Update the Target Network:
        Periodically copy the weights of the behavior network \( Q_i \) to the target network \( \hat{Q}_i \) to maintain stability in learning.
\end{itemize}

\noindent Step 4: Iterative Repetition
\begin{itemize}
    \item The algorithm repeatedly executes Steps 2 and 3 over multiple cycles, progressively optimizing each robot's decision-making and communication strategies.
\end{itemize}

\noindent Through this approach, each robot autonomously learns and adjusts its behavior based on its own experiences and interactions with other robots, thereby achieving effective task execution and information sharing in a decentralized environment. This method, which combines DQN with a dynamic communication strategy, optimizes the utilization of local information and the overall collaborative efficiency of the system.


\begin{figure}[htbp]
\vspace{4pt}
\centerline{\includegraphics[width=0.4\textwidth]{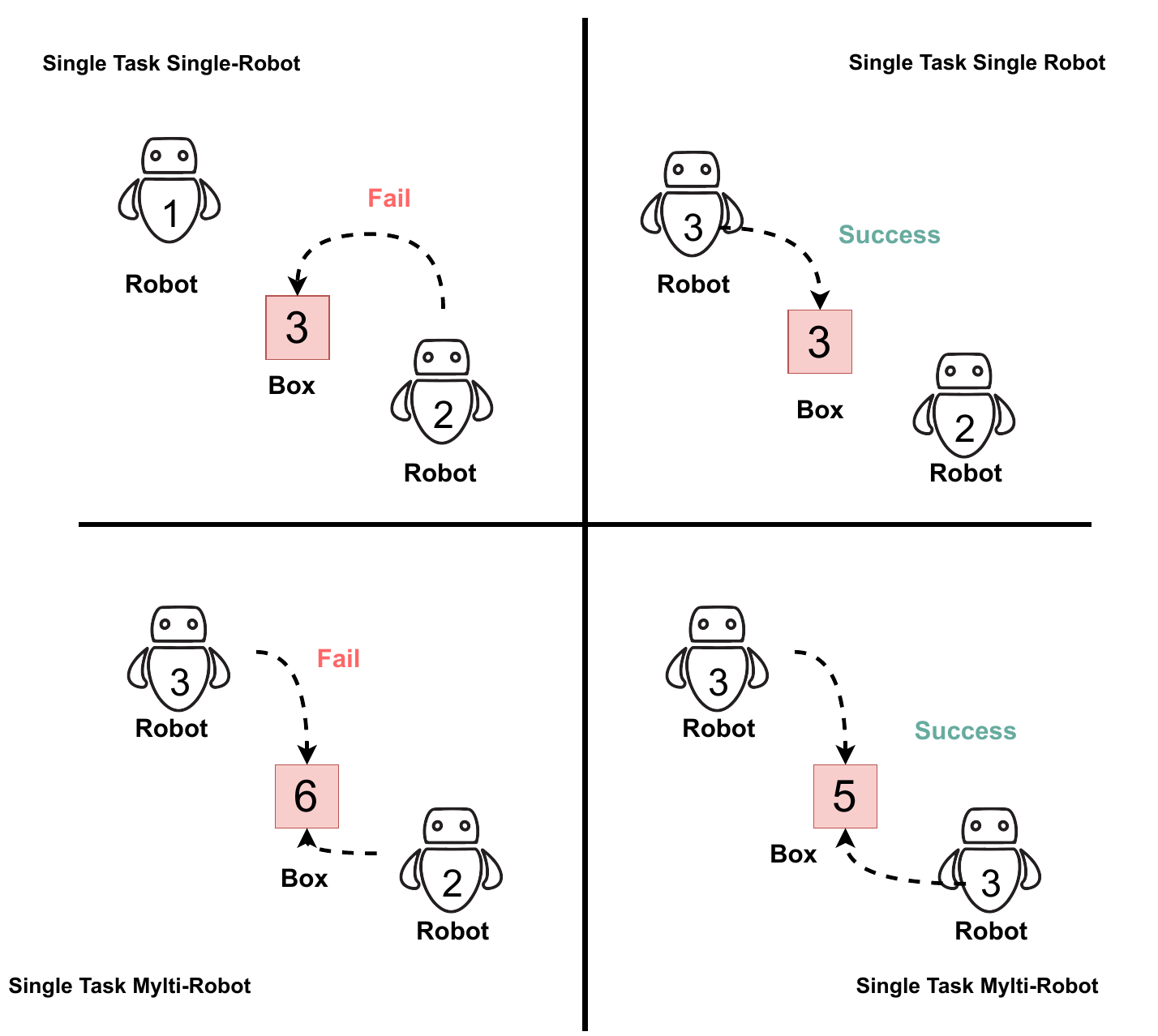}}
\captionsetup{font=footnotesize}

\caption{Task Type and Action Execution}
\label{fig:task_type}
\end{figure}

\section{Validations}
In this section, we will preliminarily validate the decentralized controller framework we previously proposed in a customized environment.

\subsection{Simulation Environment}
In the experimental environment, we have a group of \( N \) robots operating in a 10x10 grid world, where each robot has a defined load capacity, which is the maximum weight it can lift. Each box has a specific weight, which affects whether it can be moved by a specific robot or a combination of robots. There are two types of tasks, see Figure \ref{fig:task_type}:
1. Single Robot, Single Task: In this mode, one robot is responsible for finding and attempting to carry a box. If the robot's load capacity is sufficient to move the box, then the task (or subtask) is considered complete.
2. Multi-Robot, Single Task: In this mode, multiple robots collaborate to move the same box. The subtask is only considered complete when the total load capacity of all involved robots exceeds the weight of the box. The position of the box in the grid world is random, requiring the robots to search and locate the box.
Task completion criteria are categorized into two types:
Sub-task Completion: The moment one or more robots successfully lift the box, the subtask is considered complete.
Main-task Completion: The main task is only considered complete when a set number of boxes have been successfully lifted.

This experimental setting simulates real-world scenarios in multi-robot systems where robots must efficiently allocate and coordinate tasks to achieve a common goal. The setup also involves dynamic task allocation and collaboration strategies among multiple robots. This framework provides a controlled environment to develop and test algorithms for distributed decision-making and cooperative problem-solving, crucial for advancing autonomous robotic systems in practical applications.

\subsection{Environment Settings}

In this setup, our objective is to experimentally validate and compare the performance of decentralized controllers based on two types of deep reinforcement learning approaches: the standard Deep DQN and a DQN embedded with a communication strategy network.
DQN is a basic deep reinforcement learning model that does not include any additional communication mechanisms.
CE-DQN integrates a communication layer into the basic DQN framework, allowing robots to share information (such as location, status, or observed box weight) during task execution.
The comparison will focus on two main aspects:
1. Learning Curve Comparison, evaluate and compare the differences in learning efficiency during the training process between the two methods.
2. Task Completion Time Comparison, measure the efficiency of the two methods in performing actual tasks, specifically, the time required to complete tasks.
The experiment involves two types of robots:
Standard Robots, capable of completing designated load tasks.
Disturbance Robots, unable to complete the designated load tasks.
We involve four different compositions of robot teams as shown in the table \ref{tab:robot_teams}:

\begin{table}[h]
\centering
\begin{tabular}{lccc}
\textbf{Team} & \textbf{Standard} & \textbf{Disturbance} & \textbf{Total} \\
Team 1                   & 6                                  & 0                                    & 6                     \\
Team 2                    & 5                                  & 1                                    & 6                     \\
Team 3                    & 4                                  & 2                                    & 6                     \\
Team 4                    & 3                                  & 3                                    & 6                     \\
\end{tabular}
\caption{Robot Team Compositions}
\label{tab:robot_teams}
\end{table}

\subsection{Results Analysis}
\begin{figure}[htbp]
\vspace{4pt}
\centerline{\includegraphics[width=0.4\textwidth]{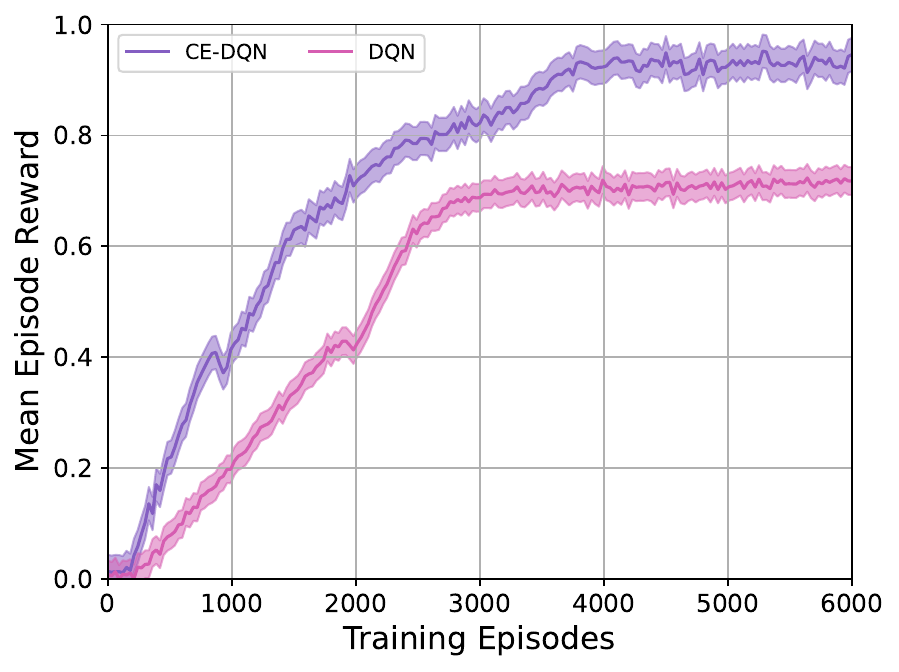}}
\captionsetup{font=footnotesize}

\caption{Learning Curve Comparison}
\label{fig: learning curve comparison}
\end{figure}

First, we compare the learning efficiency of our proposed CE-DQN algorithm and the DQN algorithm in the presence of jamming robots within a robot team. In Figure \ref{fig: learning curve comparison}, the learning curve of CE-DQN rises faster than that of DQN in the early stages of training (approximately 0 to 1000 training cycles), indicating that CE-DQN may be more effective initially and can adapt to the environment more quickly. The two curves become smoother and show less fluctuation near 6000 training cycles, with CE-DQN's curve being smoother than that of DQN, suggesting that CE-DQN's performance is more stable during training, with better adaptability and consistency in learning. In the later stages of training, the average reward of CE-DQN is higher than that of DQN, indicating that under the same conditions, CE-DQN can achieve a better performance level. Overall, the CE-DQN algorithm shows superior learning efficiency, stability, and final performance compared to the standard DQN algorithm.

\begin{figure}[htbp]
\vspace{4pt}
\centerline{\includegraphics[width=0.4\textwidth]{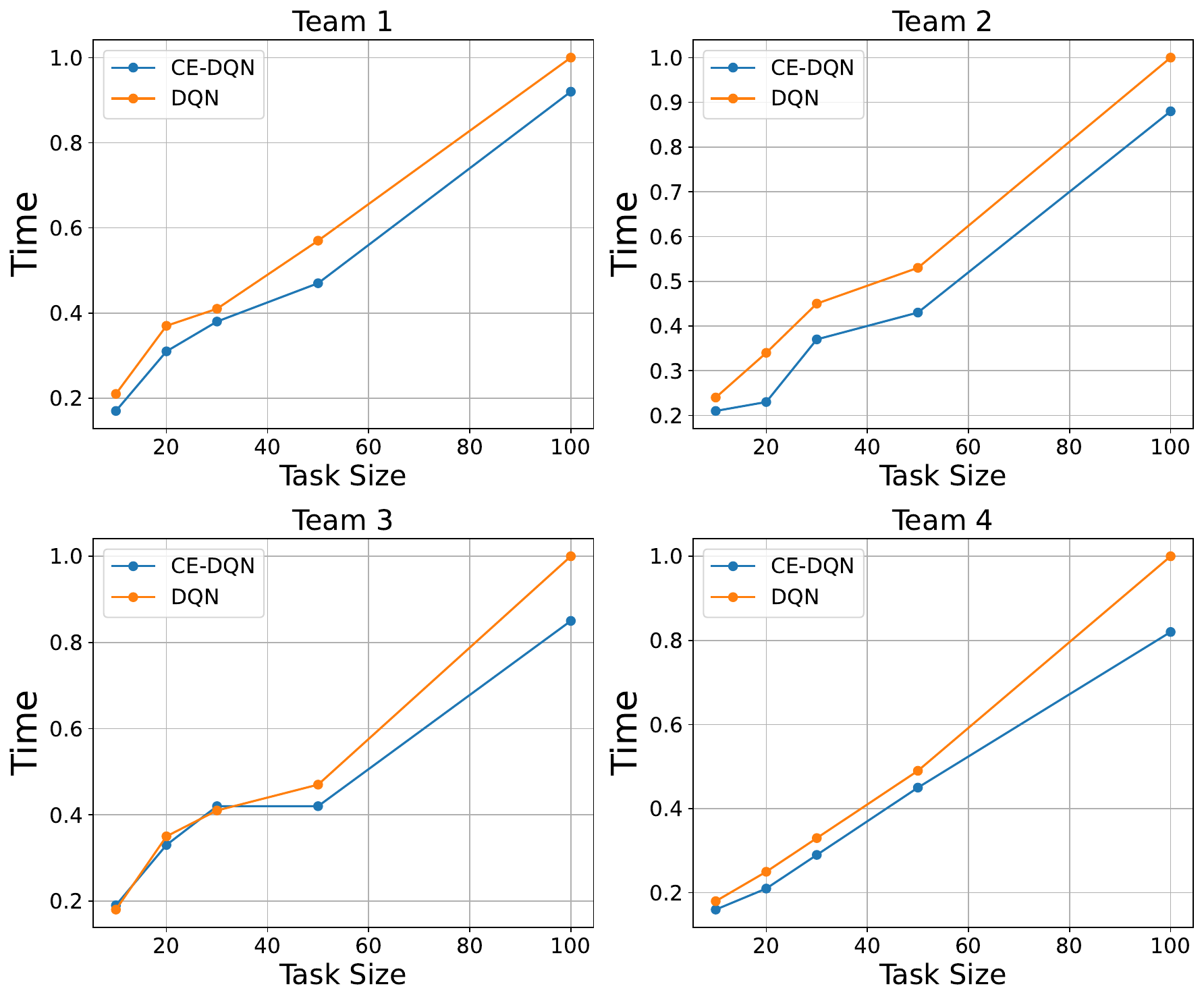}}
\captionsetup{font=footnotesize}

\caption{Task Completion Time Comparison}
\label{fig: completion time camparison}
\end{figure}

Then we compared the completion time required by the CE-DQN and DQN algorithms for handling tasks of different sizes in four different compositions of robot teams. The team compositions ranged from all normal robots to a mix of normal and disturbance robots, as shown in Figure \ref{fig: completion time camparison}.

In Team 1, for smaller tasks (size 20), the completion times of CE-DQN and DQN were close, but as the task size increased, CE-DQN showed a clear advantage over DQN in completion time, especially when the task size reached 100, where the gap was the largest.
In Team 2, as the task size increased, CE-DQN also demonstrated better performance than DQN, although the difference was smaller than in Team 1. This might be due to the presence of jamming making task completion more challenging, yet CE-DQN still maintained better efficiency.
In Team 3, CE-DQN's performance advantage became more pronounced, particularly at larger task sizes. This indicates that CE-DQN is better at adapting and optimizing resource allocation in complex environments compared to DQN.
In Team 4, in environments with more interference, CE-DQN exhibited significant performance advantages, especially at larger task sizes, suggesting that CE-DQN is more effective in handling high complexity and uncertainty.
From these results, it is evident that the CE-DQN algorithm generally performs better than the DQN algorithm across different team compositions and task sizes. This may be related to CE-DQN's decision-making process considering more and more complex factors, enabling it to better optimize strategies and decisions when facing interference and complex tasks. Additionally, CE-DQN's stability and performance advantages across all teams and task sizes indicate it may have better generalization capabilities and the ability to handle complex dynamic environments. These characteristics make CE-DQN of significant practical value in real applications, particularly in tasks that require coordination among robots of varying capabilities.
\section{Conclusion}
This study enhances the overall performance of multi-robot systems in complex environments by embedding communication strategies into the DQN framework to more effectively integrate local information. Our results clearly show that in experiments with different robot team configurations and task sizes, CE-DQN not only adapts faster and more efficient but also maintains its advantages as environmental complexity and interference level increase. Particularly in scenarios with high levels of interference, CE-DQN demonstrates significant advantages, underscoring its robustness and adaptability—qualities that are crucial for real-world application in dynamic and unpredictable environment.
Moreover, the integration of communication strategies within the DQN framework has proven to be key in achieving these improvements. By facilitating better information sharing among robots, the algorithm enhances the collective decision-making process, thereby optimizing task allocation and execution efficiency.
This research paves the way for future exploration of further improvements in decentralized control systems. Subsequent work could focus on refining communication protocols within multi-robot systems or expanding the application of CE-DQN to other domains, such as networks of autonomous vehicles or swarm robotics in exploration tasks.
In summary, the CE-DQN algorithm represents a significant advancement in the development of decentralized control mechanisms for multi-robot systems, offering a robust, efficient, and adaptable solution.





\end{document}